\documentclass[sigconf]{acmart}

\usepackage{booktabs}
\usepackage{enumitem}
\usepackage[ruled,linesnumbered]{algorithm2e}
\usepackage{subfig}
\usepackage{natbib}
\usepackage{xcolor}
\usepackage{balance}
\newcommand{\NAME}{{\em{QARAT\ }}}

\begin{document}

\title{Learning to Focus when Ranking Answers\\ \vspace{0.5cm}}

\author{Dana Sagi}
\affiliation{
  \institution{Technion --- Israel Institute of Technology}
}
\email{danasagi111@gmail.com}

\author{Tzoof Avny}
\affiliation{
  \institution{Technion --- Israel Institute of Technology}
}
\email{tzoofavny@gmail.com}

\author{Kira Radinsky}
\affiliation{
  \institution{Technion --- Israel Institute of Technology}
}
\email{kirar@cs.technion.ac.il}

\author{Eugene Agichtein}
\affiliation{
  \institution{Emory University}
}
\email{eugene.agichtein@emory.edu}

\vspace{0.5cm}
\begin{abstract}
% One of the main challenges in ranking is embedding the query and document pairs into a joint
% feature space that can then be fed to a learning-to-rank algorithm. It is common to perform intensive feature engineering that encode the similarity in the pair to achieve this representation. Recently, deep-learning solutions have shown that this representation can be achieved by only learning it from data and reaching superior results in ranking. However, those models perform poorly on longer texts or on texts with a significant part of irrelevant information or which are grammatically incorrect. To overcome this limitation, we present the use of attention mechanisms which help the model learn on which words and phrases to focus on when building the mutual representation. We show superior results on several real-world question-answer ranking datasets and provide visualization of the attention mechanism that provides more insight to how those mechanisms work.
%Replaced by Eugene
One of the main challenges in ranking is embedding the query and document pairs into a joint feature space, which can then be fed to a learning-to-rank algorithm. To achieve this representation, the conventional state of the art approaches perform extensive feature engineering that encode the similarity of the query-answer pair. Recently, deep-learning solutions have shown that it is possible to achieve comparable performance, in some settings, by learning the similarity representation directly from data. Unfortunately, previous models perform poorly on longer texts, or on texts with significant portion of irrelevant information, or which are grammatically incorrect. To overcome these limitations, we propose a novel ranking algorithm for question answering, \NAME, which uses an attention mechanism to learn on which words and phrases to focus when building the mutual representation. We demonstrate superior ranking performance on several real-world question-answer ranking datasets, and provide visualization of the attention mechanism to offer more insights into how our models of attention could benefit ranking for difficult question answering challenges.
\end{abstract}

\keywords{question answering, answer ranking, deep learning for ranking}

\makeatletter
\def\@mkbibcitation{\relax}
\makeatother

\makeatletter
\def\@copyrightpermission{\relax}
\makeatother

\makeatletter
\def\@copyrightowner{\relax}
\makeatother

\maketitle

\section{Introduction}
One of the main challenges in ranking is representing a query and document pair in a joint
feature space, which can then be fed to a ranking algorithm. Over the last decade, supervised learning-to-rank (LTR) approaches have been shown to perform best for many difficult ranking tasks, including Question Answering. However, state of the art conventional LTR approaches, such as \cite{Surdeanu:2011:LRA}, require extensive feature engineering, which includes different similarity metrics, such as manually curated lexical, syntactic and semantic similarities between the query and document.

Recently, deep learning models have obtained significant success in ranking for question answering.
However, so far, these models were shown to be successful only in modeling relatively short query and answer pairs, roughly a sentence in length \cite{severyn2013automatic}. One of the reasons these models under-perform for longer texts, is that longer answers often contain irrelevant information, which is also incorporated into the similarity representation. In this work, we explore the use of the attention mechanism for answer ranking, to overcome this limitation.

Attention mechanisms in deep learning were originally inspired by the human visual attention mechanism, which helps us perceive large amounts of information at once by focusing on parts of the information.
For example, when looking at an image, the mechanism allows us to focus on some part of it in high detail while putting less attention to the rest. 
%Encoding all information about long queries and documents into a single vector has the same limitation as looking at a large image.  
Similarly, for ranking, attention mechanism do not try to encode the entire query and document pair into a fixed-length vector, but rather
learn the interdependence between the two of them while focusing on only the important words.
%Intuitively, we want our ranking model to learn on which words to focus.
While in theory, deep-learning architectures, such as LSTM, are designed to deal with long-range dependencies -- in practice, they show poor results on representing and matching long texts \cite{LiuQCWH:2015:EMNLP}.

In this work, we present \NAME (Question-Answering Ranking with Attention), a novel deep-learning ranking algorithm for question answering, which employs attention mechanisms to identify the main question and answer terms to use for the joint representation.
%that leverages distributional sentence models to map query-document pairs to an embedding that encodes the interaction between the two while learning to identify the main interaction to focus on.
We evaluate \NAME on two popular retrieval tasks: TREC answer selection, and answer ranking for LiveQA, and show superior results on both. We complement the main results with an analysis of performance with respect to answer length, and empirically show that indeed \NAME provides significantly superior results on longer answers, and even short answers that contain  irrelevant information or are grammatically incorrect.

\begin{figure}[h!]
    \caption{Attention layer weights for a sample of questions.} 
\centering
  \includegraphics[width=0.3\textwidth]{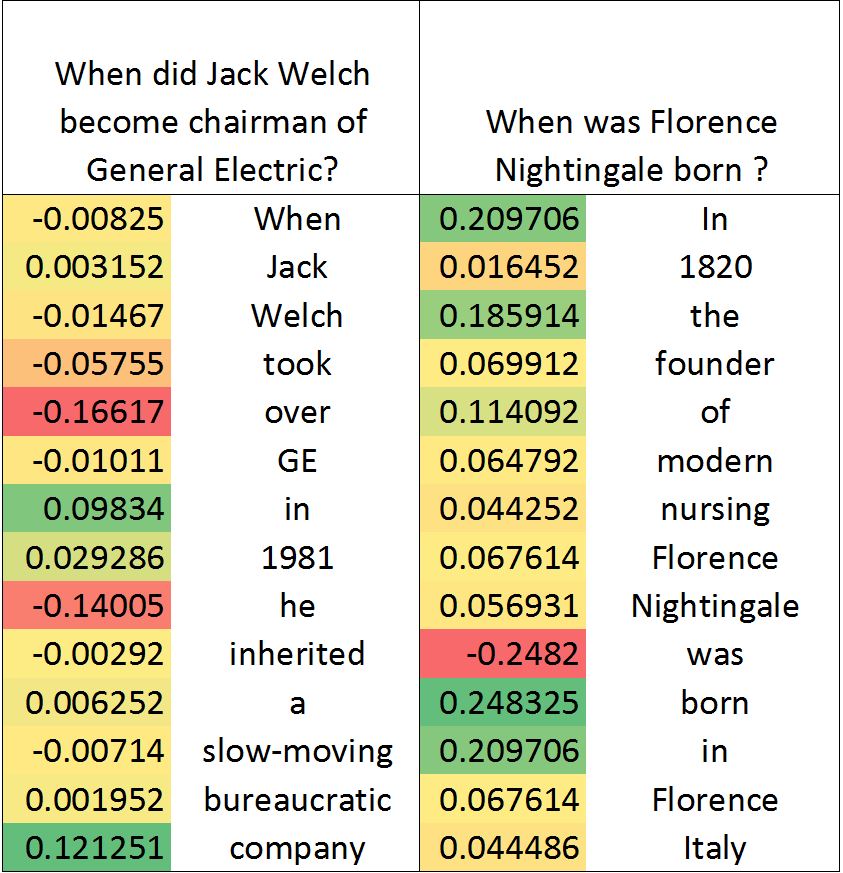}
  \label{fig:7}
\end{figure}

As an added benefit, 
%One of the criticism against deep learning methods is their lack of interpretability. 
%Attention, on the other hand, 
modeling attention explicitly enables us to visualize what \NAME learns. 
%To further understand the attention mechanism 
we report visualizations of the focus of the algorithm when embedding a pair of query and document.
Figure \ref{fig:7} shows one such visualization, where the most important words for modeling of the interdependence between the question and the answer terms are marked in green. 
In the example on the left, the phrase ``in 1981'' and  ``company'' are learned to be similar to the phrase ``chairman'' in the query, and therefore receive higher attention for ranking. Similarly, in the example on the right, the phrase ``born in'' and ``Florence'' are represented most prominently in the embedding due to their similarity to the query
%, and the word ``In'' at the beginning of the phrase as it is related to the word ``When'' of the question.

In summary, the contributions of this work are threefold: First, we propose \NAME, a novel method that improves answer ranking for long question-answer pairs via attention mechanisms. Second, we present visualizations and interpretability of our method to help gain insights into the method. Finally, we present strong empirical performance of our method on several challenging question answering datasets for answer ranking.

\section{Related Work}
Question-answer selection and ranking has been an active area of research for decades, presenting many solutions. For example, reference
\citet{wang2007jeopardy}, modeled question-answers relations to a parse tree in a way that questions and their relevant answers are connected via syntactic transformations. 
Others, e.g., \cite{heilman2010tree,severyn2013automatic} focused on 
parse tree editing for a question-answer pair, searching for minimal tree edit operations using heuristics, probabilities, and also automatic creation of features. 
Deep learning solutions have also been proposed for this task.
 \citet{yu2014deep} created a deep learning model and learned to match questions and answers by using their semantic structure. 
Other works learned the question and answer representation and matched them by similarity metric. For example, \citet{severyn2015learning} created a convolutional neural network (CNN) that receives as input vectors of question and answer pairs and returns a score for each pair. They manually added word indicator features for whether a word appeared both in the question and answer.
\citet{tan2015lstm} created an LSTM-based model by creating a biLSTM network with three gates (input, forget and output) separately for the question and the answer. They then used cosine similarity for the score analysis. This model was presented with CNN filters as well.

Attention mechanisms have been applied in several domains, including image tasks \cite{Denil:2011:corr,Larochelle:2010:nips} and other natural language processing tasks, such as machine translation \cite{BahdanauCB:2014:CoRR}, textual entailment \cite{yin2015abcnn}, etc. 
However, to the best of our knowledge, our work is the first to use attention mechanisms for the task of ranking question and answers.
Our model is based on a feed forward network with an attention layer that aims to focus on the relevant parts of the question and the answer and overcome weaknesses of other models with dealing with long or confusing answers.

\section{Question Answering Ranking with Attention}
In this section, we provide the problem definition of ranking answers for a question and present \NAME, our deep learning model with the attention mechanism.

\subsection{Problem Definition}
Given a question $q$ and possible answers for this question $a_i$ , we aim to rank candidate answers by their relevance to $q$. 
We adopt a common pointwise method for ranking.
The method requires training a binary classifier based on training instances composed of tuples of the form $(\phi(q, a_i) , y_i )$, where $y_i$ is a label indicating whether the answer $a_i$ is relevant for $q$ and $\phi$ is a function mapping the query-answer pair to a feature vector. In this work, we mainly focus on finding the best $\phi$ to represent the query-answer pairs.

% One that stands for the part in the document which is relevant to a specific query. The goal is that such a representation will have the highest similarity (defined by the ranking function) to the query representation and will get the highest score in comparison to other documents or other representation of the same document.

Most previous approaches focused on manually defining $\phi$. In this work, we present a deep learning method that learns $\phi$ from data.
This method has been explored before for related tasks, and showed promising results on short texts \cite{severyn2015learning}. Building on these efforts, we explore a new architecture and attention mechanism to learn $\phi$, which we show performs more robustly than the previous models. 

\subsection{Attention Model for Answer Ranking}
\begin{figure}[t!]
  \caption{Diagram of the \NAME \ feed forward network.}
  \centering
  \includegraphics[width=0.55\textwidth]{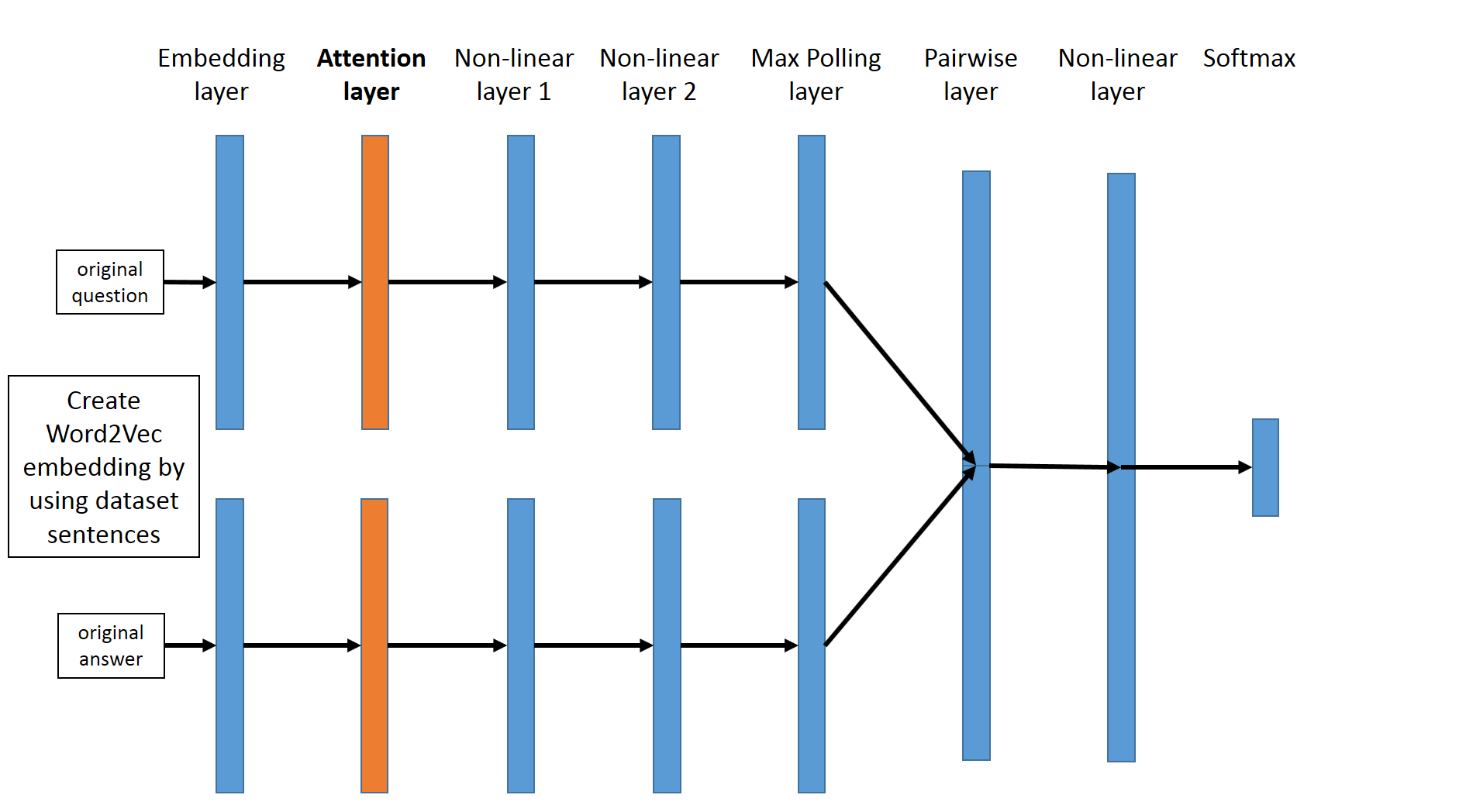}
  \label{fig:2}
\end{figure}
An attention model allows the network to sequentially focus on a subset of the input, process it, and then change its focus to another part of the input. This method makes it easier to process the data sequentially, even if the data isn't sequential in nature. In our case, the model scans the answer sequentially in order to embed it relatively to the query.

In general terms, our attention model computes a context vector for each ``time stamp'' of the sequence. The context vector is a weighted mean of the sequence states. More specifically, each state receives a weight by applying a SoftMax function over the output of an activation unit on the state. The sum of each weight,  multiplied by its state value, creates the context vector. This context vector will serve as the embedding of the question and answer.

To build $\phi$, we propose a feed forward network (shown in Figure \ref{fig:2}) that utilizes an attention layer. The inputs for the model are sentences, which represent an answer or a question. 
The network is comprised of several layers, where the first layers perform separate transformations on the question and answer, and then embed them into a single representation which is then fed to layers which output the probability of the answer responding to the input question.

%When processing the candidate answers, 
%each sentence is comprised of a list of tokens. 
%each input sentence is fixed to the length of the longest sentence in the data by padding shorter sentences with a default word. 
%The input document is divided into groups of sentences creating a matrix, with one sentence in each line. %The matrix is used as the input for the feed forward network. The output from the network is a probability %for each question-answer pair.

Specifically, the first five layers work separately on the questions and answers and subsequently merge into one layer: 
\begin{enumerate}
\item An embedding layer: replaces each token of the sentence with its word2vec representation. The Word2Vec model was trained before on the same training data which is the input for our model (i.e. it was not built on the test data). 
In addition, as suggested by \citet{severyn2015learning}, the layer concatenates to the representation a set of boolean features that represent words that appear both in the question and answer.

\item An attention layer, based on the model described by \citet{bahdanau2014neural} and \citet{raffel2015feed}. 
The layer creates the context vector as follows:
\begin{align}
h=xW+b,\\
a=Lrelu(h),\\
b=softmax(a)\\
c=\sum{b_{t}h_{t}}
\end{align}
where $x$ and $b$ are the sentence and the bias, $W$ is the parameters matrix, and $t$ is a word iterator over the sentence. $Lrelu$ is an activation function which is based on $relu$ and is defined as: 
\begin{align*}
Lrelu(x)=max\{x, 0.01x\}
\end{align*}
 The layer is further illustrated in Figure \ref{fig:3}.
\begin{figure}[H]
  \caption{Diagram of the \NAME \  attention layer}
  \centering
  \includegraphics[width=0.4\textwidth]{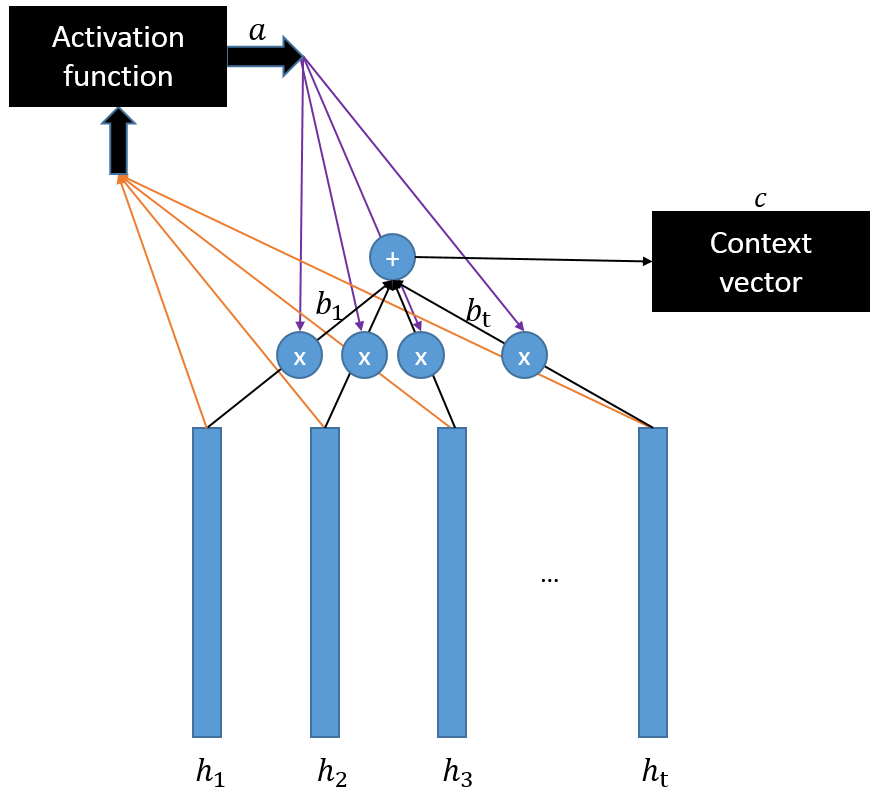}
  \label{fig:3}
\end{figure}

\item Non linearity layer, with an tanh-based activation function 
\begin{align*}
c'=tanh(cW+b)
\end{align*}
\item Non linearity layer, with an activation Lrelu function: 
\begin{align*}
c''=Lrelu(c'W+b).
\end{align*}
\item Pooling layer used to reduce the representation: max pool was used, taking the maximum activation value.
\end{enumerate}
The next layers includes the outputs from the question and answer layers, and are based on part of the network suggested by \citet{severyn2015learning}:
\begin{enumerate}
\item A pairwise layer: takes an answer and a question output vectors from the previous layers and concatenates them to a single vector.
\item A non-linearity layer, with an activation $tanh$ function.
\item A softmax layer, used in order to get a score for the question-answer pair.
\end{enumerate}

\section{Empirical Evaluation}
\subsection{Experimental Setup}
In order to test our model, we report results on two datasets:
\begin{enumerate}
\item TREC-QA answer sentence selection dataset \cite{wang2007jeopardy} contains 53,417 question-answer pairs (1,229 unique questions) from the entire TREC 8-12 collection and comes with an automatic judgment tool based on manual judgment (and not regular expressions). 
\item LIVE-QA 2015 dataset \cite{agichtein2015overview}, which contains 22,227 question-answer pairs (1187 valid questions). This dataset is characterized by both long and verbose answers, which are often not grammatically correct, and therefore might be challenging for many models.
\end{enumerate}

% The TREC-QA includes a Boolean label for each answer to mark its relevancy. The LIVE-QA dataset defines a similar task with a different scale\footnote{https://sites.google.com/site/trecliveqa2016/liveqa-qrels-2015/}, as each answer is graded from '1' (poor) to '4' (excellent), or '-2' (for non-readable answers) alternatively. In order to fit the second dataset to the same judgment tool that was used in the first one, we decided that answers with score greater then '2' will get a score of '1', otherwise they will get a score of '0'. 

Based on a separate development set, we perform parameter tuning and set 
the batch size to be 50, the em{\em lrelu} parameter to 0.01002, and the learning rate to 0.1, with parameter initialization of $W$ with 
$\sqrt{6/|a_{max}|}$, where $|a_{max}|$ is the size of the longest answer. Due to the different nature of the datasets, we set the vector size to 300 for TREC-QA and 100 for LiveQA.

% \section{Results}
% In this section, we present our empirical results and discuss the performance of the attention mechanism over several metrics.
\subsection{Main Results}
Tables \ref{tab:trec} and \ref{tab:live} summarize our results on the TREC-QA and LiveQA datasets respectively. 
We compare our model to the state-of-the-art model \cite{severyn2015learning} using the standard MRR and NDCG metrics. 
As the TREC-QA provides only binary labels, we calculated only the MRR measurement for that dataset. The results show that \NAME outperforms the state of the art on all metrics. Statistically significant results are shown in bold.
% We will present our results on each mentioned dataset using MRR and NDCG evaluation model (NDCG calculated via Kaggle method).
%\footnote{https://www.kaggle.com/wiki/NormalizedDiscountedCumulativeGain}  
\begin{table}
	\centering
\begin{tabular}{ |c|c| } 
 \hline
  & MRR \\ 
 \hline
 \citet{severyn2015learning} & 0.81 \\ 
 \hline
 Our model & \textbf{0.82} \\
 \hline
\end{tabular}
\caption{MRR Results on TREC-QA}
	\label{tab:trec}
\end{table}

\begin{table}
	\centering
\begin{tabular}{ |c|c|c| } 
 \hline
  & MRR & NDCG \\ 
 \hline
 \citet{severyn2015learning} & 0.46 & 0.7974 \\ 
 \hline
\NAME & \textbf{0.48} & \textbf{0.8018} \\
 \hline
\end{tabular}
\caption{MRR and NDCG Results on LiveQA}
	\label{tab:live}
\end{table}

\subsection{Effect of Answer and Question Length}
\begin{figure}[H]
  \caption{TREC-QA MRR results, for varying answer length}
  \centering
  \includegraphics[width=0.35\textwidth]{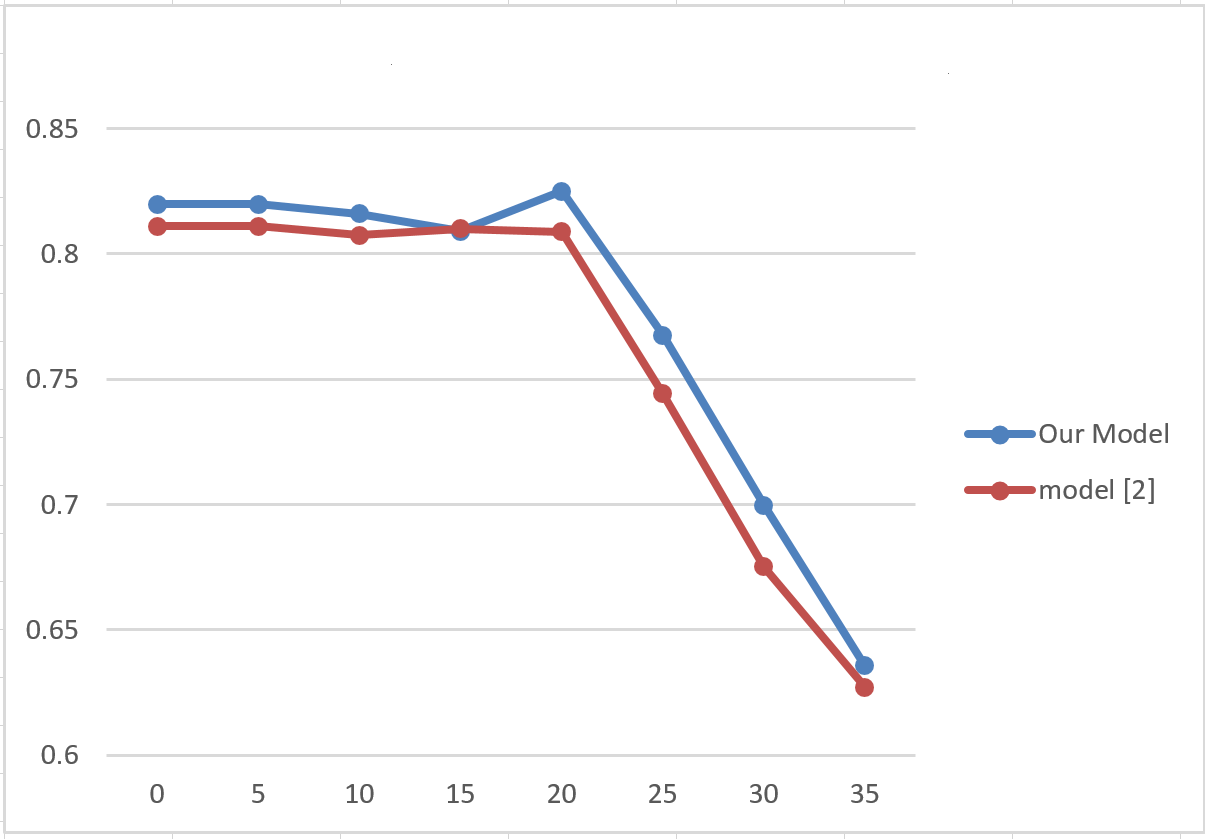}
    \label{fig:4}
\end{figure}

\begin{figure}[H]
  \caption{LiveQA MRR results, for varying answer length}
  \centering
  \includegraphics[width=0.35\textwidth]{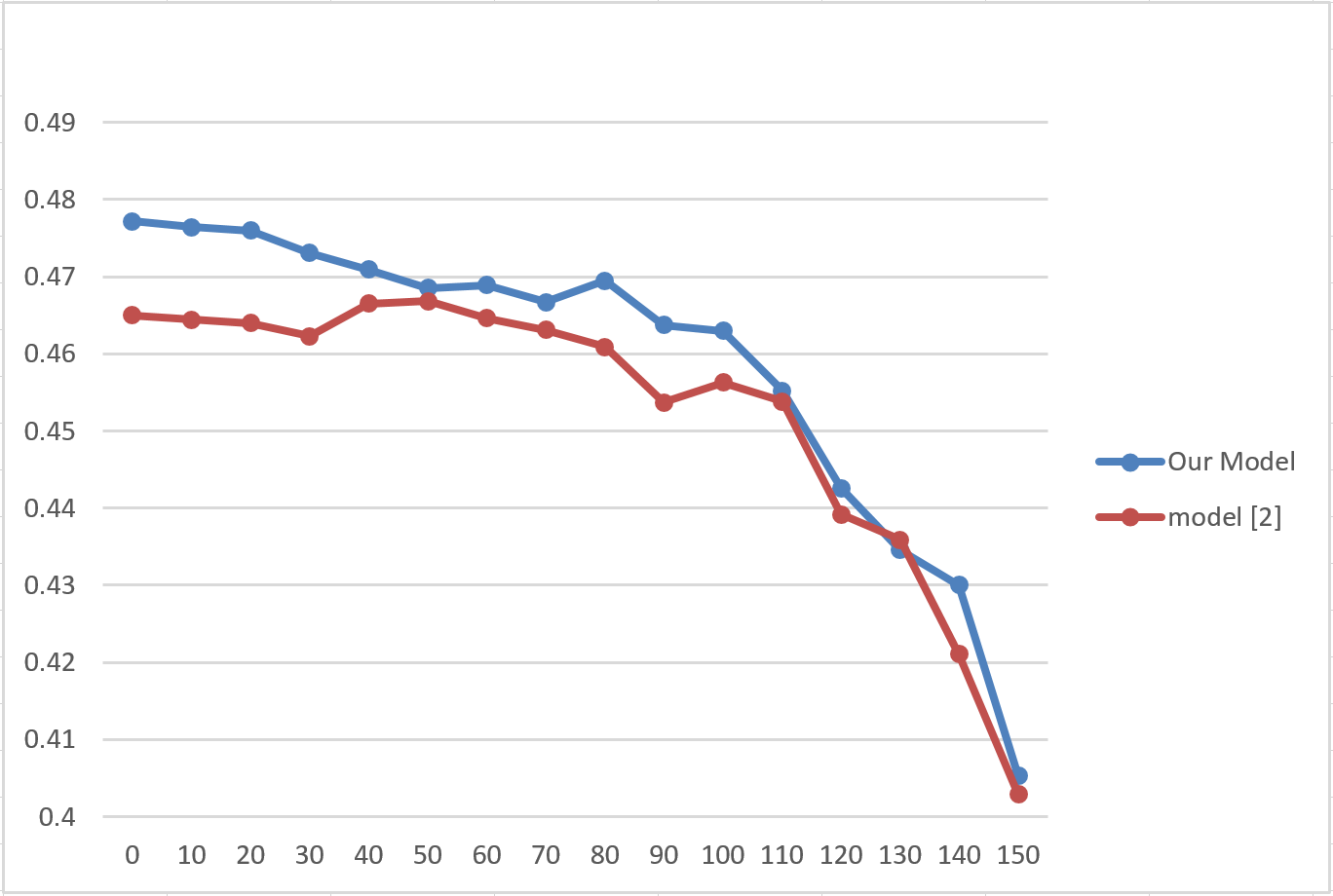}
  \label{fig:5}
\end{figure}

\begin{figure}[H]
  \caption{LiveQA NDCG results, for varying answer length}
  \centering
  \includegraphics[width=0.35\textwidth]{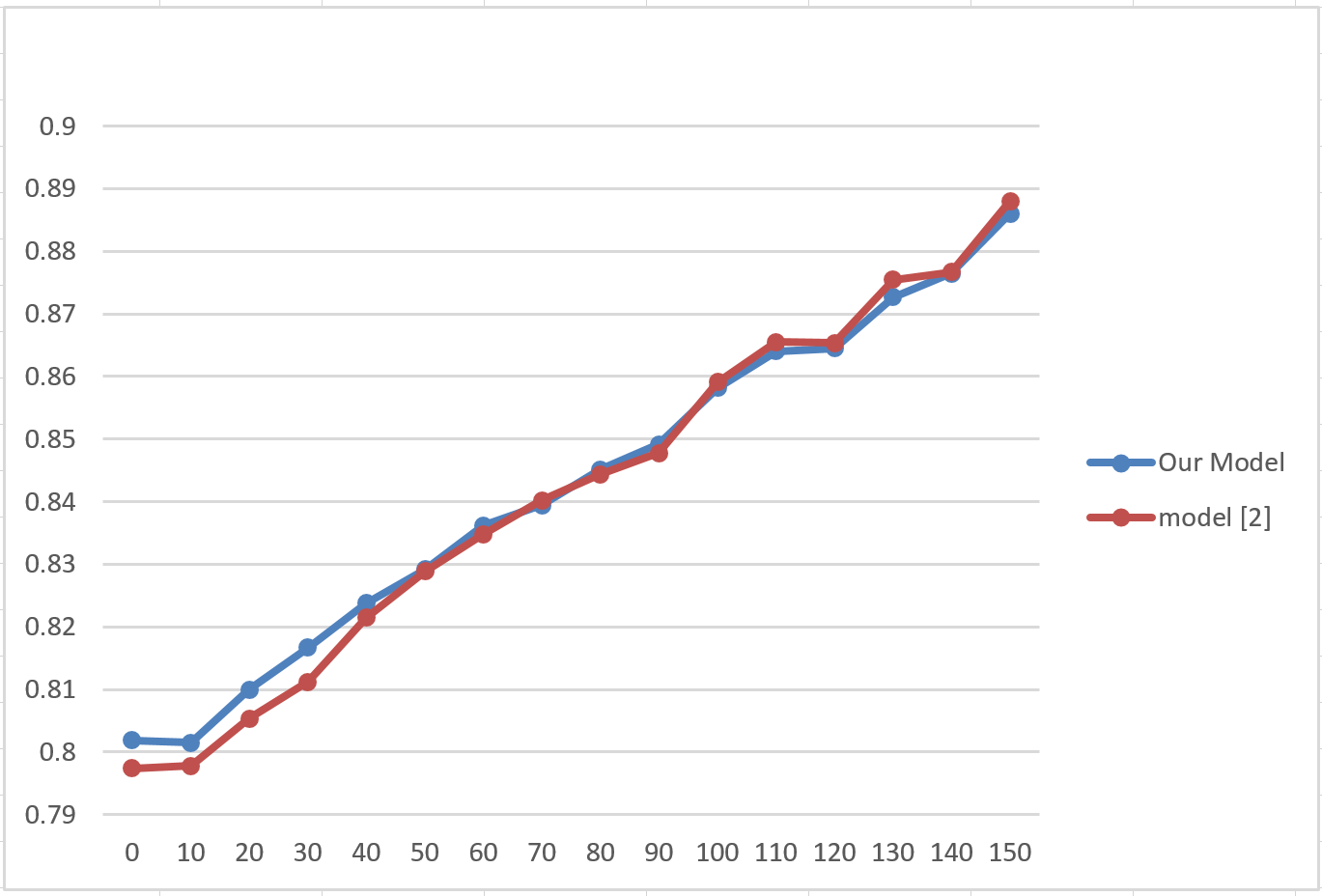}
  \label{fig:6}
\end{figure}

To further understand when our algorithm outperforms the state of the art, we compared the two models for different answers length.
Figure \ref{fig:4} shows the model results over the TREC-QA.
It is evident that the model outperforms the baseline in a statistically significant manner at answer sizes above 25.
This aligns with the strength of attention mechanisms to deal with longer texts.
When looking at LiveQA dataset (figures \ref{fig:5} and \ref{fig:6}), the model presents statistically significant results mainly for all length of answers, and specifically for those above 110 and below 30.
When investigating the answers of length less than 30, we observe that those, unlike the answers in TREC-QA, contain many phrases which are grammatically incorrect. We conclude that attention mechanism bring value for either long or even short confusing texts.

% \begin{figure}[h!]
%   \caption{TREC-QA NDCG results by answer length}
%   \centering
%   \includegraphics[width=0.5\textwidth]{2.PNG}
% \end{figure}

\subsection{Visualization of the Model}
In figure \ref{fig:7}, we present an example of two questions and their respective answers. The attention weights $W$ that \NAME created for each answer are presented to visualize where the algorithm ``attended'' the most when performing the embedding of the query-answer pair. Interestingly, the  most relevant part of the answer received the highest ``attention'' weight. In the left example, observe the phrases ``in 1981'' and ``company'' receive the highest attention weights, probably due to their relevance to the ``chairman'' phrase in the query. Our model gave this correct answer a score of 0.93 whereas the baseline gave the answer a score of 0.001. The main reason for this low score is the abundance of additional irrelevant information that influenced the baseline to score this answer low.
A similar phenomenon occurs in the example on the right, where the phrase ``born in Florence'' receives the highest attention weights when performing the embedding of the query and the answer. Our model gave this correct answer a score of 0.93 whereas the baseline gave the answer a score of 0.66.

\balance
\section{Conclusions and Future Work}
In this work, we presented \NAME, a deep learning ranking algorithms with attention mechanisms for answer ranking, and showed its superiority over the state of the art deep learning methods. Specifically, we observed that \NAME performs significantly better on longer, confusing texts with abundance of information.
To build a better intuition into the model performance, we visualized the attention mechanism to show
on which words and phrases the model focuses.
We believe the use of our proposed model will help to advance question answering research, and aid in adaption of deep learning models for ranking.
\bibliographystyle{ACM-Reference-Format}
\bibliography{references}

\end{document}